\documentclass[a4paper]{svproc}
\usepackage[T1]{fontenc}
\usepackage{graphicx}

\usepackage{multicol}
\usepackage{footmisc}
\usepackage{amsmath}
\usepackage{amsfonts}
\usepackage{tikz}
\usetikzlibrary{shapes}
\usepackage{algpseudocode}
\usepackage{algorithm}
\usepackage{cleveref}
\usepackage{subcaption}
\usepackage{url}
\usepackage{marvosym}
\usepackage[numbers]{natbib}
\begin{document}
\mainmatter
\title{Streaming Gaussian Dirichlet Random Fields for Spatial Predictions of High Dimensional Categorical Observations}
\titlerunning{Streaming Gaussian Dirichlet Random Fields}
\author{John~E.~San Soucie\inst{1,2}(\Letter)\and
Heidi~M.~Sosik\inst{2} \and
Yogesh~Girdhar\inst{2}
}

\authorrunning{San~Soucie et al.}
\tocauthor{John E. San Soucie, Heidi M. Sosik, and Yogesh Girdhar}
\institute{Massachusetts Institute of Technology, Cambridge MA, USA \and
Woods Hole Oceanographic Institution, Woods Hole MA, USA
\email{\{jsansoucie,hsosik,ygirdhar\}@whoi.edu}\\
}

\maketitle
\begin{abstract}
We present the Streaming Gaussian Dirichlet Random Field (S-GDRF) model, a novel approach for modeling a stream of spatiotemporally distributed, sparse, high-dimensional categorical observations. The proposed approach  efficiently learns global and local patterns in spatiotemporal data, allowing for fast inference and querying with a bounded time complexity. Using a high-resolution data series of plankton images classified with a neural network, we demonstrate the ability of the approach to make more accurate predictions compared to a Variational Gaussian Process (VGP), and to learn a predictive distribution of observations from streaming categorical data. S-GDRFs open the door to enabling efficient informative path planning over high-dimensional categorical observations, which until now has not been feasible. 
\end{abstract}
\section{Introduction and Background}

\subsection{Introduction}
Recent advances in in-situ sensing and autonomy have drastically improved the resolving power available for field and marine robotics and lowered deployment and operation costs. In-situ imaging systems can have temporal resolutions ranging from milliseconds to minutes, while continuous sampling on autonomous \cite{olson2007submersible} or towed \cite{cowen2008situ} platforms allows for resolving sub-meter-scale phenomena on missions lasting from hours to days. These systems produce many orders of magnitude more data than their predecessors. 

Marine autonomy poses a unique set of challenges as compared to typical terrestrial autonomy. With radio-frequency communications attenuated by water, marine autonomous systems must be capable of a high degree of autonomy without continuous access to GPS localization. Additionally, harsh environments, payload restrictions, and energy constraints limit the kinds and lengths of missions available for scientific data collection. These and other limitations incentivize the use of adaptive sampling techniques that allow scientists to collect more relevant data with the same time and energy budgets.

Adaptive sampling requires a model for how data are distributed in the world. For example, for the scalar field maximum-seeking \cite{srinivas2012information}, a Gaussian process (GP) can model the spatial distribution of the scalar field and allow for querying and prediction. Additionally, in settings with communications constraints, that model must be trained from streaming data entirely on-board the autonomous platform.  Low cost and low energy computers have been increasingly deployed on underwater vehicles. However, development of streaming statistical models for data from in-situ marine imaging systems has lagged behind the rapid progress made in imaging and computing hardware. Thus in-situ sensors generally collect data following brute-force, pre-planned missions.

Building on the Gaussian Dirichlet Random Field (GDRF) model \cite{sansoucie2020gaussian}, a previously introduced probabilistic model for spatiotemporally distributed, sparse, high-dimensional categorical data, here we present a streaming inference approach that we call streaming GDRF (S-GDRF). The S-GDRF model builds on the success of existing topic models in capturing the latent co-occurrence patterns in sparse, high-dimensional categorical data, while enabling inference and querying through the entire world. Our main contribution is a novel subsampling approach to doubly-stochastic black-box variational inference, which enables fully streaming data processing that is bounded in time and linear in space for a large category of missions.

\subsection{Background} 

\paragraph{Gaussian processes.} GPs are a natural choice of belief model for field robotics, especially when performing adaptive sampling. They can naturally encode uncertainty in complex, dynamic field conditions \cite{duckworth2021time}, and prediction allows for online Bayesian optimization approaches to path planning \cite{flaspohler2019information,rothfuss2023meta}. High-dimensional categorical observations, such as the output of a neural network classifying camera images, pose a challenge when applying existing Bayesian optimization techniques to adaptive sampling. Many adaptive sampling strategies compute expected reward rollouts only over scalar observation fields \cite{chen2022informative}, or at best low-dimensional categorical fields \cite{gilhuly2022looking}. 

Topic models, a type of Bayesian graphical model, represent the distribution of categorical observations by factoring the distributions with latent or unobserved ``topics''. The Latent Dirichlet Allocation (LDA) model captures topics with semantic meaning, organized by co-occurring clusters of words, in collections of text documents \cite{blei2003latent}. The Real-time Online Spatiotemporal Topic (ROST) model generalizes the LDA model to spatiotemporally distributed categorical observations, with a fast inference procedure suitable for embedded computation \cite{girdhar2014autonomous,girdhar2015gibbs}. The ROST model has been used to represent distributions of corals and seafloor types from robotic surveys of coral reefs \cite{jamieson2021multi}, and topics learned from a ROST model have been previously shown to capture meaningful co-occurrence relationships from phytoplankton observation data \cite{kalmbach2017phytoplankton}. 

The GDRF model, a combination of GPs and topic models into a single unified framework, was originally developed to model spatiotemporally distributed categorical data (e.g. plankton image classifications) while allowing for interpolation (e.g. time series with gaps, or path planning applications) \cite{sansoucie2020gaussian}. The original GDRF  used a mixed, offline inference algorithm. Existing streaming variational inference algorithms are generally limited to specific prior-likelihood combinations, such as exponential family models \cite{broderick2013streaming,mcinerny2015population}. In the context of streaming Markov Chain Monte Carlo inference, different approaches to subsampling from past data have been shown to impact convergence properties of the model \cite{girdhar2014unsupervised}. Here we use a similar subsampling approach, but replace the Markov Chain Monte Carlo algorithm with a black-box variational inference algorithm to learn an approximate posterior \cite{ranganath2014black}. 

Between GPs, topic models, and their unification in GDRFs, none provides a suitable belief model for adaptive sampling over high-dimensional categorical data. This work seeks to fill this gap --- the lack of predictive models necessary for computing expected reward rollouts in informative path planning tasks --- by providing a belief model for spatiotemporally distributed, sparse, high-dimensional categorical observations.

\section{Methods}
\subsection{Technical Approach}

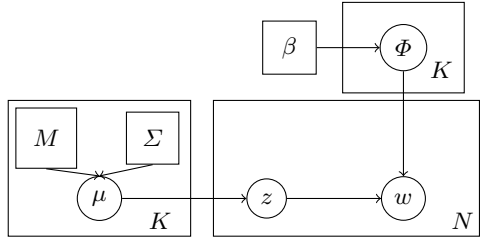
\begin{figure}[htb]
    \centering
    \begin{tikzpicture}

\node (mu) at (0,0) [circle, draw] {$\mu$};
\node (z) at (2.2,0) [circle, draw] {$z$};
\node (w) at (4,0) [circle, draw] {$w$};
\node (mean) at (-.7,.8) [regular polygon,regular polygon sides=4, draw] {$M$};
\node (covariance) at (.7,.8) [regular polygon,regular polygon sides=4, draw] {$\Sigma$};

\node (psis) at (4,2) [circle, draw] {$\Phi$};
\node (b) at (2.5,2) [regular polygon,regular polygon sides=4, draw] {$\beta$};

\draw[->] (b.east) -- (psis.west);

\draw[->] (mean.south) -- (mu.north);
\draw[->] (covariance.south) -- (mu.north);

\draw[->] (mu.east) -- (z.west);
\draw[->] (z.east) -- (w.west);
\draw[->] (psis.south) -- (w.north);

\draw (-1.2,-.5) rectangle (1.2,1.3);
\draw (1.5,-.5) rectangle (5,1.3);
\draw (3.2,1.4) rectangle (4.8,2.6);

\node at (.8,-.3) {$K$};
\node at (4.78,-.3) {$N$};
\node at (4.5,1.7) {$K$};
\end{tikzpicture}
    \caption{The graphical model for GDRFs (reproduced from \cite{sansoucie2020gaussian})}
    \label{fig:graphical_GDRF}
\end{figure}

\paragraph{Gaussian-Dirichlet Random Fields.} 
At a high level, a GDRF is a smooth, continuous model for categorical data which factors observations into latent co-occurrence patterns. A GDRF models the distribution of categorical data (``observations’’) in a spatiotemporal world by introducing latent communities, each of which represents a distribution over observation types. The relationship between communities and observations is modeled with Dirichlet distributions, while the distribution of communities in space-time is modeled by GPs. Formally, a GDRF is a continuous Bayesian model defined on a (potentially infinite) set of points $\mathbf{X} = \{\mathbf{x}_1, \mathbf{x}_2, \dots\}$, each of which we call \textit{locations} in the \textit{world} $\mathbf{X}$. \textit{Observations} $w_i$ and \textit{communities} $z_i$ are $W$- and $K$-categorical variables, respectively. The community spatial distributions are computed by passing a set of GPs $\mu_j$, representing mean latent log probability fields for each community through a \textit{link function} $f_j: \mathbb{R}^K \to \left[0, 1\right]$, where $\sum_jf_j(\mu_1, \dots, \mu_K) = 1$, e.g. the \textit{softmax} function. The Dirichlet-distributed global community distributions $\Phi$ are multiplied by the spatial probabilities to get the observation distributions (\cref{fig:graphical_GDRF}).

\paragraph{Streaming inference for S-GDRFs.} S-GDRFs augment the GDRF model with an inference procedure capable of handling streaming observations, as well as a sparse inducing-point approximation to allow for computationally tractable streaming. The GDRF prior and likelihood functions do not permit a simple conjugate posterior, so inference of the posterior to a given prior $p(x, z)$ on observations $x$ and latent variables $z$ must be approximate. We utilize stochastic black-box variational inference (BBVI) \cite{ranganath2014black}, with a key alteration to the final step of selecting observations to calculate the loss $\mathcal{L}$. After sampling latent variables from the approximate posterior, BBVI involves calculating the log probability of the observations. For the S-GDRF model, we allow the inference step to choose $N_s$ points from any of the previously observed data, with a custom probability distribution $\pi_{t}^{N_s}(x)$ that can be weighted to bias random selection of recent observations. Training iterations can occur between observation of new data points, and the number of such iterations can vary if desired. The sparse inducing point approximation\cite{titsias2009variational} reduces the complexity of covariance matrix inversions from $O(n^3)$ to $O(nm^2)$ with $m$ inducing points. The selection of a finite collection of $n_s$ past observations for calculating the loss allows the upper bound to be fixed at $O(n_sm^3)$ prior to training, even on an infinite stream of data. 

\paragraph{Evaluation of GDRF predictive power} We compare the S-GDRF model to a single-GP-per-category regression model, trained entirely offline, which we refer to as the Variational Gaussian Process (VGP) model. To evaluate the predictive power of each model, we define the predictive Kullbach-Liebler divergence (PKL) metric: for a model trained on the first $N$ data points in a stream, we query the observation distribution at all future locations to be observed. The PKL metric $M_{PKL}$ for a single future observation is the KL divergence between the predicted and observed distribution of observation categories: 
\begin{equation}
    M_{PKL}\left(p_i\left(w_{i+1:}|x_{i+1:}\right)\right) = D_{KL}\left(p_i\left(w_{i+1:}, x_{i+1:}\right) ||p_{\text{emp}}\left(w_{i+1:}, x_{i+1:}\right)\right) 
    \label{eq:mpkl}
\end{equation}
where $p_i$ is a model trained on the first $i$ observations, $w_{i+1:}$ are the following observations, $x_{i+1:}$ are their locations, and $p_{\text{emp}}$ is the empirical distribution of observations.

\subsection{Experiments}

\begin{figure}[tb]
    \centering
    \includegraphics[width=\textwidth]{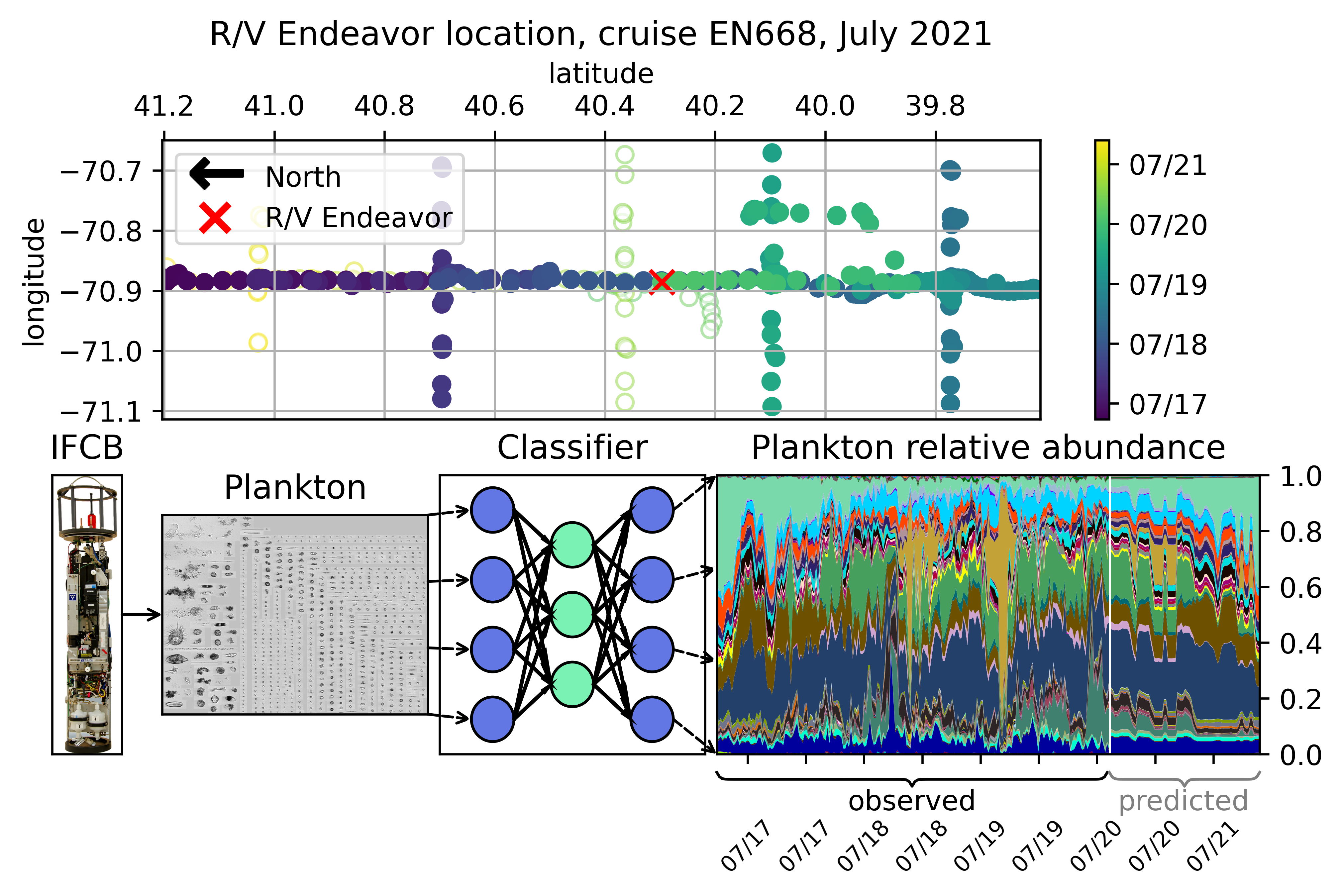}
    \caption{In July 2021 aboard the R/V Endeavor, an Imaging FlowCytobot was used to take high-throughput, high-resolution images of plankton in surface seawater. Post-cruise, all images were classified with a neural network-based classifier. The S-GDRF model enables prediction of this kind of spatiotemporally distributed, sparse, high-dimensional categorical data.}
    \label{fig:ifcb_cnn_gdrf_schematic}
\end{figure}

\paragraph{Temporal data.} We compared the SGDRF and VGP approaches using a dataset collected on a July 2021 research cruise on the \textit{R/V Endeavor} (EN688) and classified via neural network (\cref{fig:ifcb_cnn_gdrf_schematic}). As part of the Northeast U.S. Shelf Long-Term Ecological Research (NES-LTER) project, the cruise sampled along an established north/south transect ranging from Martha’s Vineyard in the north to the continental slope beyond the Mid-Atlantic Bight in the south, punctuated by latitudinally sparse segments with high longitudinal resolution (\cref{fig:ifcb_cnn_gdrf_schematic}). An Imaging FlowCytobot (IFCB) \cite{olson2007submersible}, an automated imaging-in-flow cytometer, was connected to the ship’s underway seawater system and used to collect thousands of images of microscopic plankton in surface seawater, approximately once every 25 minutes. After the cruise, a pre-trained convolutional neural network classifier sorted the images into over 150 different classes \cite{orenstein2015whoi}.  We trained both an S-GDRF model and a series of VGP models on this data in a streaming fashion, in order to evaluate their predictive power.

\begin{figure}[htb]
\centering
\begin{subfigure}[b]{0.48\textwidth}
\centering
\includegraphics[width=0.8\textwidth]{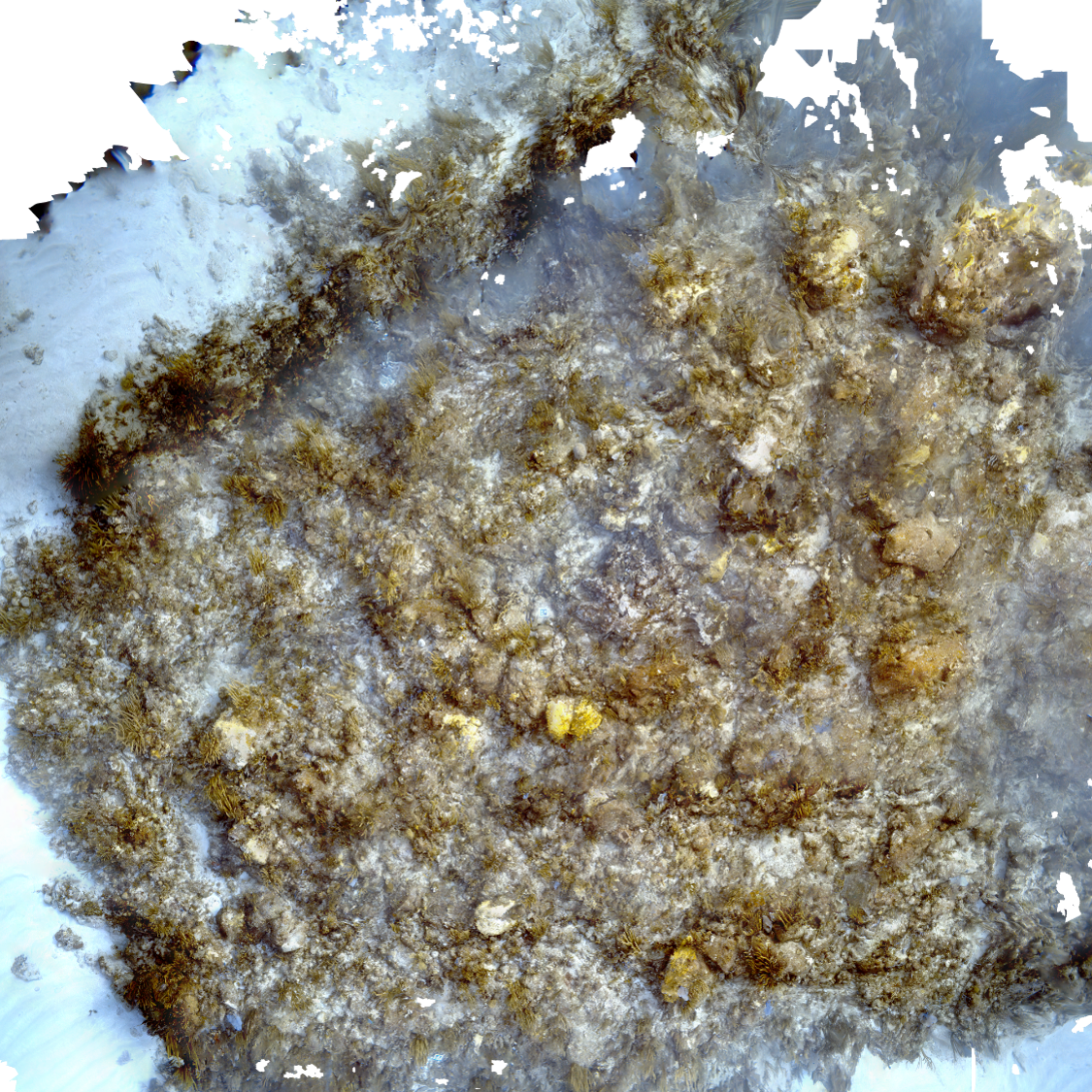}
\caption{Joel's Shoal reef, US Virgin Islands}
\label{fig:sub_joelshoal}
\end{subfigure}
\hfill
\begin{subfigure}[b]{0.48\textwidth}
\centering
\includegraphics[width=0.8\textwidth]{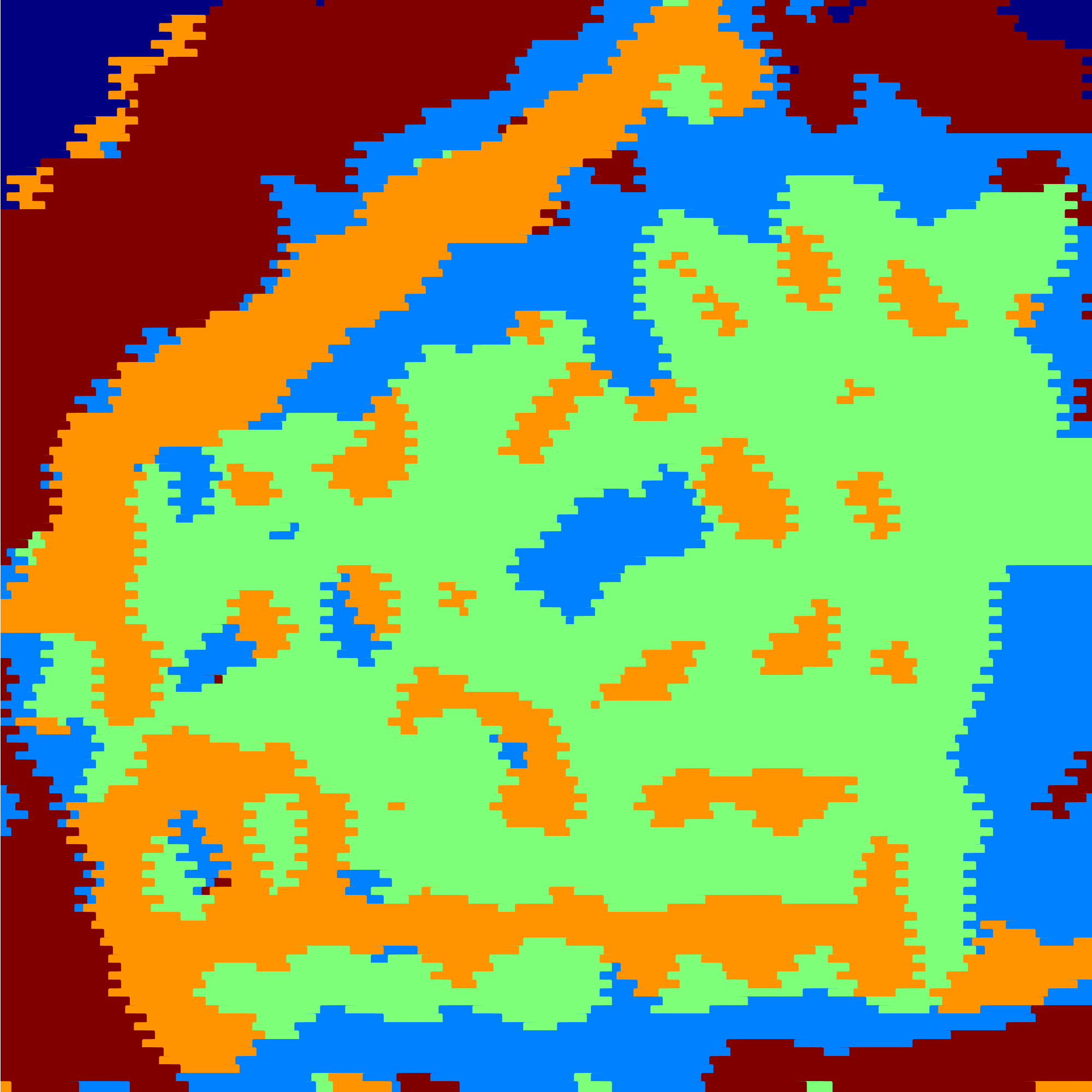}
\caption{S-GDRF max likelihood community}
\label{fig:sub_coral}
\end{subfigure}
\caption{S-GDRF inference on the Joel's Shoal coral head shows reasonable labelings for pixels in a 130x130 grid with 15436 unique extracted features. }
\label{fig:coral}
\end{figure}

\paragraph{Spatial data.} Images of Joel's Shoal Reef, St. John, US Virgin Islands were taken with an underwater robotic system in November 2022 \cite{girdhar2023curee}. These were stitched together into a single mosaic of the reef, and 15436 simple visual features \cite{rublee2011orb} were extracted in a 130x130 cell grid (\cref{fig:sub_joelshoal}). A lawnmower trajectory over this grid was simulated, and observations from this trajectory were fed into an S-GDRF model to evaluate scalability and performance on two-dimensional data (\cref{fig:sub_coral}).

\section{Results}

\label{sec:result}

\begin{figure}[htb]
    \centering
    \includegraphics[width=\textwidth]{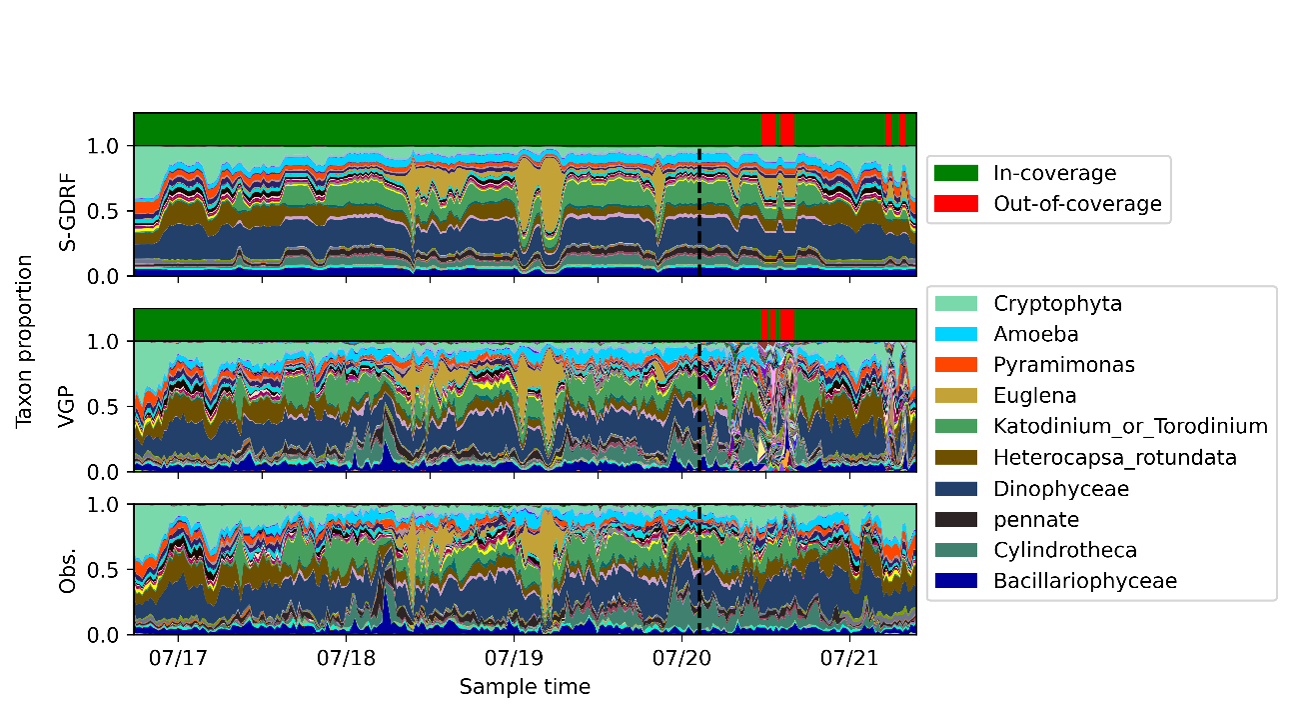}
    \caption{Time-series plots of the distribution of plankton, with model fits (left of the dashed line) and predictions (right of the dashed line) in out-of-coverage areas (more than one kernel lengthscale from any datapoint left of the black line).
    Due to the large number of plankton taxa, only the top ten taxa (by mean observed relative abundance) are marked in the legend.}
    \label{fig:obs_gdrf_vgp}
\end{figure}

\begin{figure}[htb]
    \centering
    \includegraphics[width=0.8\textwidth]{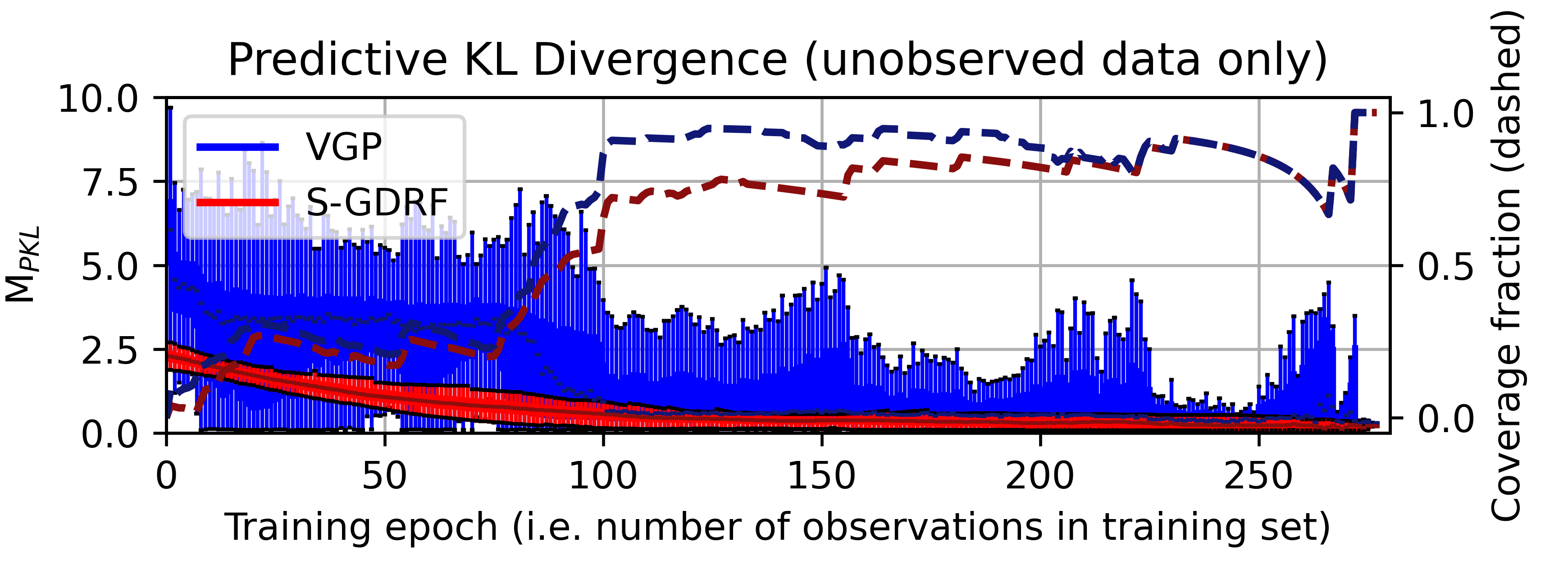}
    \caption{S-GDRF and VGP predictive Kullbach-Liebler divergence (PKL) metric, for the 1-D temporal dataset. The x-axis is the number of training data, while the y-axis represents the PKL metric, i.e. the KL divergence between model predictions on all other data and observations. The median and quartile PKL for all unobserved (i.e. non-training) data is displayed with solid lines. The dashed lines represent the coverage fraction, i.e. what proportion of the unobserved data points like within one GP kernel lengthscale of a previously observed datapoint.}
    \label{fig:boxplot}
\end{figure}

\paragraph{1-dimensional S-GDRF prediction.} A mid-training snapshot of plankton predicted and actual relative abundances highlights how predictions from S-GDRF are more useful than from VGP (\cref{fig:obs_gdrf_vgp}). The S-GDRF model tends to take a community-mean-field approach to prediction in out-of-coverage regions, while the VGP model generally makes a noisy and uninformative prediction. For observed and in-coverage data, the VGP model captures much of the high-frequency variability seen in the observations. In contrast, the S-GDRF model effectively smooths the observation distributions. Turning to predictive capability, the S-GDRF model has a consistently lower median PKL metric than the VGP model across the entire cruise, as well as lower variance (\cref{fig:boxplot}). Median VGP PKL drops rapidly as the coverage fraction increases, reaching a reasonably low value once the coverage fraction is above 0.8. However, VGP variance remains high throughout the data series.

\paragraph{2-dimensional S-GDRF inference.} After running a simulated lawnmower trajectory over the Joel's shoal reef, the final inferred maximum likelihood topics (\cref{fig:sub_coral}) visually capture the major structures in the the coral reef (\cref{fig:sub_joelshoal}). On average, training proceeded at 7.68 iterations per second on an NVIDIA A6000 GPU. 

\section{Discussion}
\paragraph{Advantages over sparse and streaming GPs.} The sparse GP\cite{snelson2005sparse}, with its data-independent inducing points, opened the door for streaming GP inference algorithms which handle new data points as they arrive in a computationally efficient manner. Many streaming GP algorithms (e.g. \cite{ranganathan2010online}) extend standard GP inference, and only work for Gaussian likelihoods. Streaming variational GP inference algorithms can use non-Gaussian likelihoods \cite{bui2017streaming}. However, these methods still require a strictly Gaussian stochastic process (as opposed to a more general Bayesian structure), and even the most efficient models have not been demonstrated to work with hundreds or thousands of categories in a GP classifier. In contrast, the S-GDRF is a factorizes the observation model and uses a GP to only model the low-dimensional latent representation of the data. Our experiments show that the S-GDRF is able to accurately capture the distribution of sparse, high-dimensional categorical data, while being extremely computationally efficient. The coral reef experiment (\cref{fig:coral}) demonstrated efficient ($0.13$ seconds per iteration) inference with over $15000$ observation categories, as S-GDRF time and memory complexity generally scale with the amount of data and the number of latent GPs. In fact, no comparison to a strictly GP-based model (such as a VGP) is computationally feasible for these data. The low computational cost of S-GDRF, albeit on workstation hardware, supports its use on the embedded computers available for in-situ inference.

\paragraph{S-GDRF outperforms VGP.} The S-GDRF model outperforms VGP according to the prediction metric, especially for out-of-coverage data. The S-GDRF latent variables likely aid in prediction here. The smaller number of GPs increases robustness to noise, while at the same time the Dirichlet distributions pool global information about relevant co-occurrence patterns. Even if no useful information exists about the distribution of communities far away from any observations, knowing the kinds of co-occurrence patterns that can mix to produce observations provides better predictions.The simulated results presented here support the use of S-GDRF models for streaming inference on autonomous platforms, and specifically as a belief model for informative path planning over categorical observations. The S-GDRF model outperformed a VGP model with more parameters and a much larger computational budget, training on the entire streaming dataset in minutes. A single refinement step, with a 25x25 uniform inducing point grid, takes less than a second on state-of-the-art hardware. This implies real-time computation is feasible on CUDA-capable embedded systems. The Appendix in the supplementary material contains a demo underscoring this, in which a 15000-category S-GDRF is trained on a 130x130 cell grid on desktop hardware.

\section{Conclusion}

The S-GDRF model, a streaming Bayesian model for spatiotemporally distributed, sparse, high-dimensional categorical data, was proposed as a belief model for robotic path planning over categorical data. This model was demonstrated to outperform a single-GP-per-observation model at streaming prediction tasks on a 1-dimensional temporal dataset. Evaluation on a 2-dimensional spatial dataset highlighted both the scalability of the S-GDRF model in terms of the number of observation categories and the computational efficiency of the S-GDRF inference algorithm. 
These results demonstrate how S-GDRFs can enable efficient informative path planning over high-dimensional categorical observations.
Future work includes deployment of the S-GDRF algorithm on in-situ robotic platforms for real-time belief modeling and path planning.

\subsubsection*{\ackname}
This project was funded in part by NSF (NRI Award No. 2133029 to YG, LTER Award No. 1655686 to HMS), the Simons Foundation (grant 561126 to HMS), and the NDSEG Fellowship Program (to JESS).
\bibliography{bibliography}

\end{document}